\documentclass{article} % For LaTeX2e
\usepackage{iclr2025_conference,times}

\usepackage{amsmath,amsfonts,bm}

\def\eqref#1{equation~\ref{#1}}
\def\1{\bm{1}}

\DeclareMathAlphabet{\mathsfit}{\encodingdefault}{\sfdefault}{m}{sl}
\SetMathAlphabet{\mathsfit}{bold}{\encodingdefault}{\sfdefault}{bx}{n}

\usepackage{hyperref}
\usepackage{graphicx}
\usepackage{amsmath}
\usepackage{booktabs} 
\usepackage{array} 
\usepackage{float}
\usepackage{hyperref}
\usepackage{changepage}
\usepackage{colortbl}
\usepackage{tabularray}
\usepackage{xcolor}
\usepackage{url}
\usepackage{color}
\usepackage{wrapfig}
\usepackage{multirow} 

\newcommand{\gray}[1]{\textcolor{gray}{#1}}

\title{3D-AffordanceLLM: Harnessing Large Language Models for Open-Vocabulary Affordance Detection in 3D Worlds}

\author{Hengshuo Chu$^1$, Xiang Deng$\dag$$^1$, Qi Lv$^1$, Xiaoyang Chen$^1$,Yinchuan Li$^2$, Jianye Hao$^2$, Liqiang Nie$\dag$$^1$\\
$^1$Harbin Institute of Technology (Shenzhen), $^2$Huawei Noah’s Ark Lab\\
}

\iclrfinalcopy % Uncomment for camera-ready version, but NOT for submission.
\begin{document}

\maketitle
\renewcommand{\thefootnote}{$\dag$} 
\footnotetext[1]{Corresponding authors}
\begin{abstract}
3D Affordance detection is a challenging problem with broad applications on various robotic tasks. 
Existing methods typically formulate the detection paradigm as a label-based semantic segmentation task.
This paradigm relies on predefined labels and lacks the ability to comprehend complex natural language, resulting in limited generalization in open-world scene.
To address these limitations, we reformulate the traditional affordance detection paradigm into \textit{Instruction Reasoning Affordance Segmentation} (IRAS) task. 
This task is designed to output a affordance mask region given a query reasoning text, which avoids fixed categories of input labels.
We accordingly propose the \textit{3D-AffordanceLLM} (3D-ADLLM), a framework designed for reasoning affordance detection in 3D open-scene.
Specifically, 3D-ADLLM introduces large language models (LLMs) to 3D affordance perception with a custom-designed decoder for generating affordance masks, thus achieving open-world reasoning affordance detection.
In addition, given the scarcity of 3D affordance datasets for training large models, we seek to extract knowledge from general segmentation data and transfer it to affordance detection.
Thus, we propose a multi-stage training strategy that begins with a novel pre-training task, i.e., \textit{Referring Object Part Segmentation}~(ROPS).
This stage is designed to equip the model with general recognition and segmentation capabilities at the object-part level.
Then followed by fine-tuning with the IRAS task, 3D-ADLLM obtains the reasoning ability for affordance detection. 
In summary, 3D-ADLLM leverages the rich world knowledge and human-object interaction reasoning ability of LLMs, achieving approximately an 8\% improvement in mIoU on open-vocabulary affordance detection tasks.

\end{abstract}
\section{Introduction}
Robots are increasingly integrating into various aspects of our daily life~\citep{matheson2019human}. 
As we progress toward developing the next generation of more advanced robotic agents, it is essential to enable robots to comprehend natural language instructions within context and to perceive task-relevant information in their surroundings.
This skill is particularly vital for seamless interactions in unstructured environments, such as homes, where adaptability to diverse situations is crucial.
Specifically, the robots need to not only identify the objects in the environments but also locate the specific regions of each object that are suitable for interaction: \textit{affordance}.

{The concept of affordance was introduced by ecological psychologist James Gibson~\citep{Gibson1966-GIBTSC-5} and has since played a significant role in various robotic applications, including object recognition~\citep{hong20233d,hou2021affordance}, action anticipation~\citep{roy2021action}, agent activity recognition~\citep{chen2023affordance}, and object functionality understanding~\citep{li2023locate}. 
Affordance describes potential interactions between robots and their environment, such as using a knife's blade for cutting tasks. Detecting affordances is challenging due to object diversity and complexity~\citep{min2016affordance}. 
Traditionally, 2D images and CNNs are 
used~\citep{nguyen2016detecting,do2018affordancenet,pacheco2022one}\citep{krizhevsky2012imagenet}, but 2D information lacks the depth necessary for precise manipulation, necessitating 3D transformations~\citep{deng20213daffordancenet}. 

{With advanced depth cameras, 3D point clouds have become a widely used modality in robotic applications ~\citep{liu2019deep}. Unlike conventional images, 3D point clouds offer robots direct and detailed 3D information about surrounding objects and environments. 
Hence, the 3D affordance detection has been deemed as a critical step in bridging perception and manipulation in the physical world for an embodied agent, thus has
shown substantial impact on practical applications such as robotic manipulation~\citep{geng2023rlafford,moldovan2012learning}.
While current approaches are limited by fixed label sets~\citep{deng20213daffordancenet,mo2022o2o}, reducing flexibility and generalization in dynamic settings. 
To overcome the fixed label set problem in affordance detection, Nguyen et al.~\citep{nguyen2023open} have incorporated a text encoder to enable models to handle certain levels of open-vocabulary detection, but these algorithms still rely on a classification based training paradigm.
As a result, they lack the ability for rapid and continuous learning when presented with new affordance label data. Furthermore, current affordance detection methods also heavily rely on the predefined labels and lack the ability to understand and reason over long contextual text.
Additionally, the scarcity of 3D affordance datasets~\citep{deng20213daffordancenet,nguyen2023open} constrains the effective training of large-scale models.

Towards these issues, we redefine the 3D affordance detection as an \textit{Instruction Reasoning Affordance Segmentation} (IRAS) task and accordingly propose \textit{3D-AffordanceLLM}~(3D-ADLLM). 
The IRAS task is designed to output an affordance mask region in response to complex, reasoning-based query text, overcoming the limitations of fixed affordance labels and the difficulty of understanding complex instructions. 
Our  3D-ADLLM framework introduces large language models (LLMs) to 3D affordance perception with a specifically designed decoder for generating affordance masks, thus achieving open-world reasoning affordance detection.
Specifically, we introduce an additional token, \texttt{<AFF>}, into the original LLM vocabulary. When the \texttt{<AFF>} token is generated, its hidden embedding is further decoded into the corresponding segmentation mask. 
By representing the segmentation mask as an embedding, 3D-ADLLM not only gains segmentation capability but also benefits from end-to-end training.
However, due to the scarcity of 3D affordance datasets for training large models, we propose a multi-stage training strategy to extract knowledge from general segmentation data and transfer it to affordance detection.
This process involves pre-training on PartNet~\citep{Mo_2019_CVPRPartnet} with \textit{Referring Object Part Segmentation} (ROPS) tasks to acquire the object-part level general recognition and segmentation knowledge.
Subsequently, we fine-tune the model with the IRAS task to achieve context-aware reasoning ability and robust performance in open-set zero-shot affordance detection.

Our main contributions are summarized as follows:
\begin{itemize}
\item 
Different from the existing affordance detection methods that rely on fixed sets of labels, we address this limitation by introducing a new detection paradigm based on the \textit{Instruction Reasoning Affordance Segmentation} (IRAS) task.
By reforming the label-based semantic segmentation task in the traditional affordance detection paradigm into a natural language-driven reasoning affordance segmentation task, our model enables more flexible and context-aware reasoning, facilitating effective zero-shot learning capabilities.

\item 
To address the IRAS tasks driven by
semantic complex natural language, we consequently propose the \textit{3D AffordanceLLM} (3D-ADLLM) model, combining a large language model (LLM) with a carefully designed Affordance Decoder.
Our 3D-ADLLM framework can understand semantically-rich, long-context instructions and leverages the LLM's world knowledge for superior open-vocabulary affordance detection.
\item 
Due to the scarcity of 3D affordance datasets for training large models, we propose a multi-stage training strategy to transfer general segmentation knowledge into affordance detection.
First, the model is equipped with general recognition and segmentation knowledge through a novel pretraining task, i.e., the Referring Object Part Segmentation (ROPS). 
Subsequently, the model is fine-tuned with the IRAS task to handle context-aware reasoning and 
affordance region prediction.
\end{itemize}

\section{Related Work}
\textbf{Affordance Detection.}
Originating from the 2D domain, initial work in affordance detection primarily focused on identifying objects with affordances~\citep{do2018affordancenet}. 
Building on this foundation, later studies~\citep{lu2022phrase} introduced linguistic descriptions to improve detection, but they continued to emphasize object-level affordances, lacking fine-grained analysis. 
Addressing this problem, subsequent research~\citep{chen2023affordance, li2023locate, luo2022learning, nagarajan2019grounded, mi2020intention} has focused on detecting specific affordance parts, establishing a new benchmark for precision in the field.
With the advancement of embodied AI, the scope of affordance learning has expanded into 3D domain.
3D AffordanceNet~\citep{deng20213daffordancenet} introduces the first benchmark dataset for learning affordance from object point clouds.
IAGNet~\citep{yang2023IAGNet} propose a setting for learning 3D affordance parts guided by image queries. 
Recently, some work~\citep{nguyen2023open} also explores the open-vocabulay affordance detection in point clouds.
However, these methods primarily focus on linking object geometric features with fixed affordance labels, overlooking the semantic aspect.
This limitation makes it challenging to understand natural language instructions and hampers the ability to generalize affordance detection to unseen scenarios. 
In contrast, the proposed 3D-ADLLM overcomes the limitations of fixed label sets and enhance the ability to comprehend semantic complex description. 
Specifically, we shift the detection paradigm from label-based semantic segmentation into Instruction Reasoning Affordance Segmentation (IRAS).

\textbf{3D Large Multi-Modal Models.}
3D object-level LMMs~\citep{yu2022pointbert,xue2023ulip,zhou2023uni3d} have successfully bridged the gap between 3D vision and text by leveraging large-scale 3D object datasets like~\citep{deitke2023objaverse,vishwanath2009modelnet}.
ShapeLLM~\citep{qi2024shapellm} further advances the embodied interaction and referring expression grounding through its novel and powerful point encoder.
However, despite these advances, such models still face challenges in interpreting complex spatial relationships within 3D scenes.
For scene-level LMMs, models like Chat-3D~\citep{wang2023chat3d} and LL3DA~\citep{chen2024ll3da} enable interaction with scene objects using pre-selection mechanisms. 
Building on this foundation, Chat-3D v2 ~\citep{huang2023chat3dv2} enhances referencing and grounding accuracy by incorporating object identifiers, while 3D-LLM ~\citep{hong20233dllm} improves scene comprehension by integrating positional embeddings and location tokens.
Unlike previous works that primarily focus on 3D grounding and understanding, our method introduces a specialized token, \texttt{<AFF>}, which enables LLMs to directly detect affordances and generate affordance masks within 3D open-world scene.

\section{Method}
\subsection{Paradigm Reformulation}
Affordance detection aims to identify specific regions of objects that are suitable for interaction.
It has been deemed as a critical step in bridging perception and manipulation in the physical world for embodied agents. 
As illustrated in Fig.~\ref{fig:IRAS task} (a), the traditional paradigm uses a shared point backbone~\citep{qi2017pointnet++,zhao2021point_transformer,wang2019dynamic} to extract point-wise features, and generates masks with a predefined type semantic segmentation head. Alternatively, they leverage a text encoder like CLIP~\citep{radford2021cliplearning} to associate point-wise features with text embeddings of affordance labels using cosine similarity, achieving limited open-vocabulary detection on the phrase level.
This paradigm relies on predefined labels and has a limited ability to understand complex natural language, which restricts its generalization in 3D open-world scene.

\begin{figure}[htbp]
    \centering 
    \includegraphics[width=1.0\linewidth]{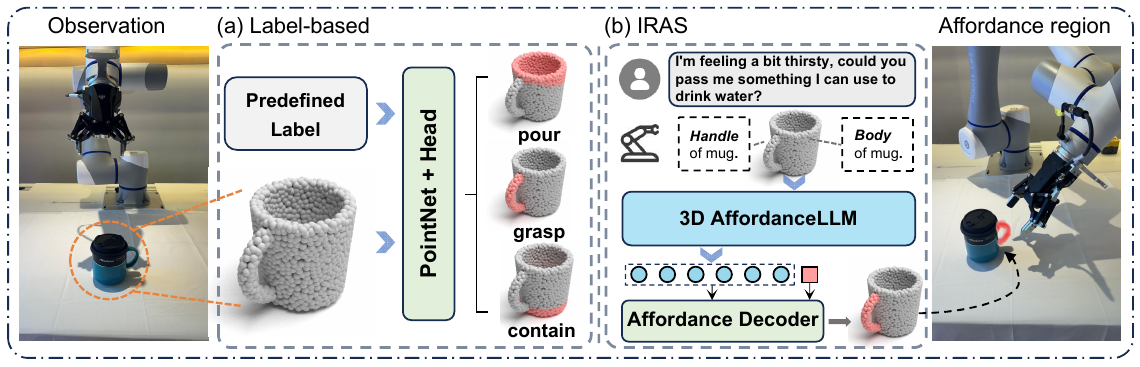}
    \vspace{-12pt}
    \caption{
    The comparison of the affordance detection paradigm based on our IRAS or traditional label-based segmentation tasks. (a) shows that label-based paradigm can only detect the fixed set of affordance regions through the predefined label and seg-head; (b) demonstrates the IRAS
     based paradigm forges a link between semantic complex instruction and object affordance, enabling open-world reasoning affordance detection. 
    }
    \vspace{-10pt}
    \label{fig:IRAS task}
\end{figure}

To address these limitations, we introduce a new paradigm formulated as an \textbf{I}nstruction \textbf{R}easoning \textbf{A}ffordance \textbf{S}egmentation (\textbf{IRAS}) task as depicted in Fig.~\ref{fig:IRAS task} (b). 
This paradigm is designed to establish a robust connection between language context and object affordance, avoiding the overreliance on auxiliary affordance label prediction. 
This approach facilitates a significant improvement in our ability to understand and interact with the physical world.

\textbf{IRAS Definition.} \textit{Given a query reasoning instruction $Q_{a}$ and an object point cloud $P_{c} \in \mathbb{R}^{n \times 3}$ with N points, the goal of IRAS is to predict a binary mask of $M_{a} \in \mathbb{R}^{N}$ that delineates the functional regions pertinent to the query, affordance regions:} 
$${F}_{Model}(Q_{a}, P_{c}) \Rightarrow M_{a}$$

\subsection{3D-AffordanceLLM}
\label{sec:model}
To the traditional methods that rely on fixed label sets and are limited to short-text detection, IRAS demands robust language comprehension and reasoning to associate the potential affordance in input query with 3D objects areas.
Thus, we incorporate large language models (LLMs) into 3D affordance perception.
LLMs, trained on trillions of tokens, excel in understanding and reasoning about instructions, and possess extensive world knowledge. 
For instance, when asked where to interact with a mug to grasp it, LLMs suggests using the handle for a firm grip to avoid spilling. 
This demonstrates LLMs' world knowledge and the capability in understanding human-object interactions. 
To harness this capability for 3D affordance perception, we introduce the 3D AffordanceLLM Model, aiming to improve affordance detection in previously unseen contexts.

Our framework, \textit{3D AffordanceLLM}, as illustrated in Fig.~\ref{model_architecture}, primarily consists of two main components:
(1) a point cloud multimodal model which is trained to accept point cloud and text inputs and generate response, including a special token, \texttt{<AFF>}; 
(2) an Affordance Decoder (AFD), which extracts hidden layer features from these \texttt{<AFF>} tokens and combines them with segmentation point features to generate affordance masks.
\begin{figure}[htbp]
    \centering    
    \includegraphics[width=1.0\linewidth]{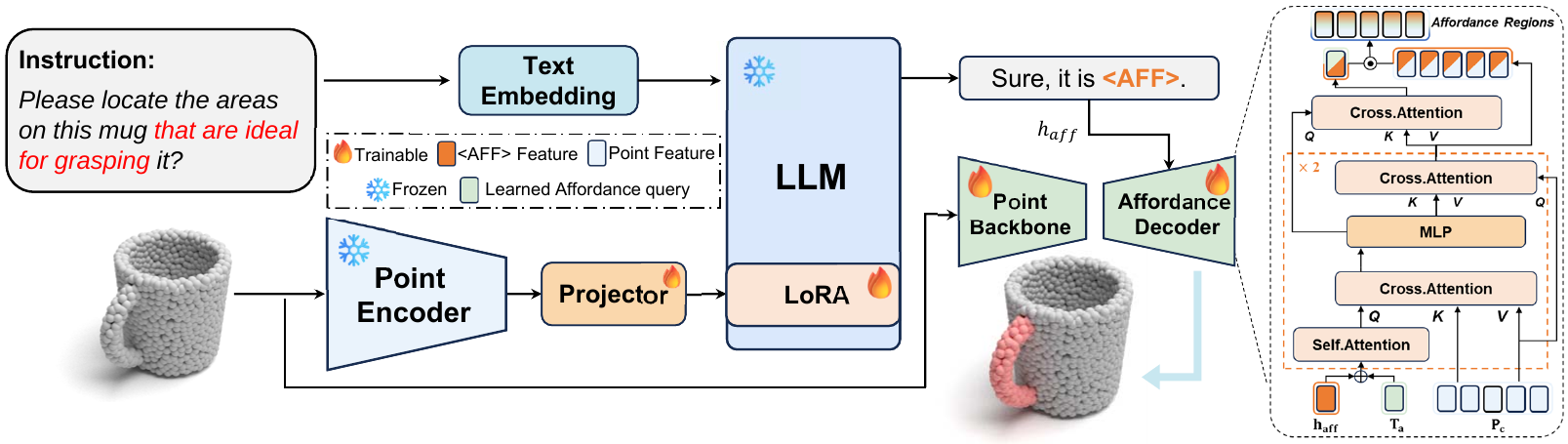}
    \vspace{-1.5em}
    \caption{The Pipeline of 3D-ADLLM. Given the input point cloud and query reasoning instruction, the point cloud multimodal model is trained with lora to predict special token \texttt{<AFF>}. Finally, the special token and dense point features from $f_\mathrm{PB}$ is fed into our designed affordance decoder to generate the final affordance mask.}
    \label{model_architecture}
    % \vspace{-5pt}
\end{figure}

\subsubsection{Model Architecture}
As is shown in Fig.~\ref{model_architecture}, our 3D AffordanceLLM consists of the following modules: a pre-trained point cloud encoder $f_\mathrm{pe}$,a projector $f_\mathrm{proj}$, a point backbone $f_\mathrm{PB}$, an affordance decoder $f_\mathrm{AFD}$ and a pre-trained large language model (LLM) backbone $f_\mathrm{llm}$.

\textbf{Point Encoder.} The point cloud encoder $f_\mathrm{pe}$, takes a point cloud $\mathbf{P}_\mathrm{cloud} \in \mathbb{R}^{n \times d}$ as input, where $n$ represents the number of points and $d$  denotes the feature dimension of each point. 
The output of the encoder is a sequence of point features $X = (x_\mathrm{1}, x_\mathrm{2}, \dots, x_\mathrm{m}) \in \mathbb{R}^{m \times c}$, where $m$ is the number of point features and $c$ is the feature dimension. 
Similarly, the point backbone $f_\mathrm{PB}$, also processes input point cloud $\mathbf{P}_\mathrm{cloud} \in \mathbb{R}^{n \times d}$, extracting
the dense point cloud features $X' = (x'_\mathrm{1}, x'_\mathrm{2}, \dots, x'_\mathrm{n}) \in \mathbb{R}^{n \times c'}$, specifically tailored for segmentation tasks. 
These features are subsequently fed into the Affordance Decoder.

\textbf{LLM Projector.} The projector $f_\mathrm{proj}$ is a MLP layer that maps the point features $X$ to point tokens $Y = (y_1, y_2, \dots, y_m) \in \mathbb{R}^{m \times c''}$, where $c''$ is the dimension of the point tokens, matching the dimension of the text tokens.

\textbf{Large Language Model.} The LLM backbone \( f_\mathrm{llm} \) is a decoder-only Transformer model~\citep{AshishVaswani16}, which processes a sequence of tokens comprising text and point tokens. 
This mixed token sequence is denoted as \( Z = (z_1, z_2, \dots, z_k) \in \mathbb{R}^{k \times c''} \), where \( k \) is the total number of tokens. Leveraging a self-attention mechanism, the LLM backbone captures contextual relationships between different token types, enabling it to generate responses based on both text and point cloud inputs. 
Formally, the output of the LLM backbone \( f_\mathrm{llm} \) is a sequence of predicted tokens \( \hat{Z} = (\hat{z}_1, \hat{z}_2, \dots, \hat{z}_k) \in \mathbb{R}^{k \times c''} \). 
The prediction of the \( i \)-th token, \( \hat{z}_i \), is conditioned on all previous tokens, \( Z_{<i} = (z_1, \dots, z_{i-1}) \), which can be expressed mathematically as:
$$
\hat{z}_i = f_\mathrm{llm}(Z_{<i}).
$$
Each \( \hat{z}_i \) is passed through a final linear layer followed by a softmax operation, which maps the hidden states to a probability distribution over the vocabulary. 
This layer is denoted as \( f_\mathrm{vocab}: \mathbb{R}^{c'} \to \mathbb{R}^V \), where \( V \) is the size of the vocabulary. The final prediction \( \tilde{z}_i \) for the \( i \)-th token is the word in the vocabulary with the highest probability, expressed as:
$$
\tilde{z}_i = \arg\max_{w \in \mathrm{vocab}} f_\mathrm{vocab}(\hat{z}_i)[w].
$$
\textbf{Affordance Decoder.}
Building on the success of learnable query-based methods in object segmentation, we introduce an Affordance Decoder Module (AFD) that leverages a set of learnable output queries conditioned on input questions, termed \textit{affordance queries} $T_\mathrm{a}$ to decode segmentation masks. 
A two-layer decoder updates both the point features and the question features via cross-attention. 
Then, the updated query tokens and point features are used to dynamically predict affordance masks.

\subsubsection{Embedding as Affordance}
Unlike conventional tasks such as grounding, question answering, etc., within the realm of 3D large multi-modal models (LMMs), the IRAS task is depicted to generate a affordance segmentation mask directly given a reasoning query.
Most current 3D LLM (such as 3D-LLM~\citep{hong20233d}, ShapeLLM~\citep{qi2024shapellm} support 3D scenes or objects and text as input, but they can only output text or bbox and cannot directly output fine-grained segmentation masks. 
Inspired by the LISA model~\citep{lai2024lisa}, which directly outputs the segmentation mask in the 2D domain, we adopt a similar idea in 3D affordance detection. 
To achieve that, we propose the embedding-as-affordance paradigm to inject new affordance segmentation capabilities into the 3D LMM. The pipeline of our method is illustrated in Fig.~\ref{model_architecture}. Specifically, we expand the original LLM vocabulary by adding a new token, \texttt{<AFF>}, which signals a request for an affordance output. 
Given a complex reasoning instruction query $\mathbf{Q}_\mathrm{aff}$ and a point cloud input $\mathbf{P}_\mathrm{cloud}$, we feed them into the multimodal point clouds LLM ${F}_\mathrm{3D-ADLLM}$, which outputs a text response $\hat{\mathbf{y}}_\mathrm{txt}$: "Sure, it is $\texttt{<AFF>}$." 
This process can be formulated as:
$$
\hat{\mathbf{y}}_\mathrm{txt} = {F}_\mathrm{3D-ADLLM}(\mathbf{P}_\mathrm{cloud}, \mathbf{Q}_\mathrm{aff}).
$$
When the LLM intends to generate a binary affordance mask, the output $\hat{\mathbf{y}}_\mathrm{txt}$ would include a \texttt{<AFF>} token. 
We then extract the LLM last-layer embedding 
$\tilde{\mathbf{h}}_\mathrm{aff}$ corresponding to the \texttt{<AFF>} token and apply an MLP projection layer $\mathrm{Proj}$ to obtain $\mathbf{h}_\mathrm{aff}$. 
Simultaneously, the point cloud backbone \( f_\mathrm{PB} \) extracts the dense point clouds features $\mathbf{f}$ from the points input $\mathbf{P}_\mathrm{cloud}$. Finally, $\mathbf{h}_\mathrm{aff}$ and $\mathbf{f}$ are fed to the decoder \( f_\mathrm{AFD}\) to produce the final affordance mask $\hat{\mathbf{M}}_\mathrm{aff}$. 
% \red{The detailed structure of the decoder \( f_{AFD}\) is simlilar to the mask decoder in SAM.} 
The process can be formulated as
$$\mathbf{h}_\mathrm{aff}=\mathrm{Proj}(\tilde{\mathbf{h}}_\mathrm{aff})$$
$$\quad\mathbf{f}=f_\mathrm{PB}(\mathbf{P}_\mathrm{cloud}),\\\hat{\mathbf{M}}_\mathrm{aff}=f_\mathrm{AFD}(\mathbf{h}_\mathrm{aff},\mathbf{f}).$$

\subsection{Multi-Stage Training}
\label{sec:two stage training strategy}
Existing 3D affordance datasets, such as 3D AffordanceNet datasets, OpenAD datasets in~\citep{deng20213daffordancenet,nguyen2023open}, are constrained in availability and dataset sizes.
Thus, given the scarcity of 3D affordance datasets for training large models, we devise a multi-stage training strategy which extracts knowledge from general segmentation data and transfers it to IRAS affordance detection.
In addition, due to the varying scales of target affordance regions,  we propose a sample unbalanced loss factor to enhance the model's learning effectiveness and adaptability across different region scales.

\begin{figure}[tbp]
    \centering    
    \includegraphics[width=1.0\linewidth]{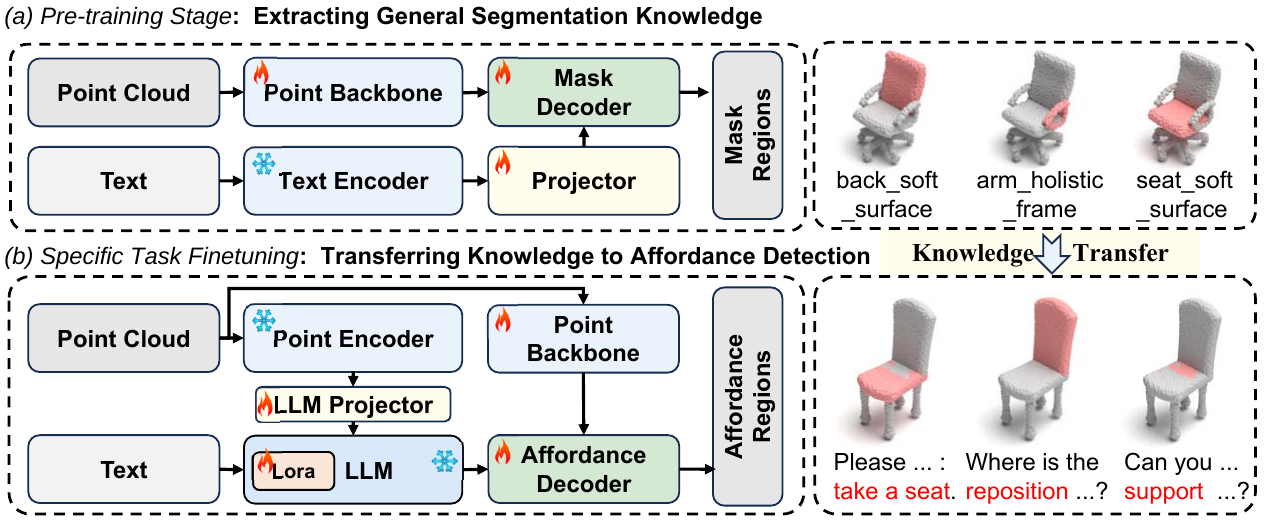}
    \vspace{-10pt}
    \caption{Multi-stage training strategy. Illustration of transferring general segmentation knowledge to affordance detection. (a) depicts the process of extracting general segmentation knowledge, while (b) illustrates the framework for transferring this knowledge to affordance detection}
    \label{Two-stage training}
    \vspace{-10pt}
\end{figure}

\subsubsection{Extracting General Segmentation Knowledge}
Considering the limited amounts of affordance datasets for training large models, this stage aims to leverage general datasets to equip the model with general recognition and segmentation capabilities at the object-part level. 
Thus, we introduce \textbf{R}eferring \textbf{O}bject \textbf{P}art \textbf{S}egmentation (\textbf{ROPS}) task to acquire the general knowledge.

\textbf{ROPS Definition.} 
\textit{Given a referring expression that denotes the name of the object's components $Q_{}$ and an object point cloud $P_{c} \in \mathbb{R}^{n \times 3}$ consisting of N points, the objective of ROPS is to predict a binary mask for $M_{p} \in \mathbb{R}^{N}$ that corresponds to the query:}
$${F}_{Model}(Q_{p}, P_{c}) \Rightarrow M_{p}$$

In the pre-training phase, we employ the framework in Fig.~\ref{Two-stage training} (a) to train the ROPS task on the PartNet dataset~\citep{Mo_2019_CVPRPartnet}. 
As depicted in Fig.~\ref{Two-stage training} (a), a trainable backbone processes the object point cloud to extract the characteristics of the point $\mathbf{f}_\mathrm{P_{cloud}}$. The descriptions of the parts of the objects are encoded using a frozen text encoder to generate text features $\mathbf{f}_\mathrm{Q_{part}}$, which are then mapped via an offline layer of MLP to produce $\mathbf{f'}_\mathrm{Q_{part}}$. 
Finally, $\mathbf{f'}_\mathrm{Q_{part}}$ and $\mathbf{f}_\mathrm{P_{cloud}}$ are passed into the Mask Decoder to generate the final part mask $\mathbf{M}_\mathrm{part}$, formulated as:
$$
\mathbf{M}_\mathrm{part} = \mathrm{Mask Decoder}(\mathbf{f'}_\mathrm{Q_{part}} , \mathbf{f}_\mathrm{P_{cloud}})
$$

\textbf{Training Objectives.}
Unlike~\citep{Mo_2019_CVPRPartnet}, which uses a multi-class head for prediction, our stragety seeks to extract the knowledge relationship between referring object part description and the corresponding mask regions.
Thus, we solely employ Dice Loss and Binary CrossEntropy (BCE) loss to guide the segmentation mask prediction.
$$\mathcal{L}=\lambda_\mathrm{1}\mathcal{L}_\mathrm{BCE}+\lambda_\mathrm{2}\mathcal{L}_\mathrm{DICE}.$$  

\subsubsection{Transferring Knowledge to Affordance Detection}

Building upon the extensive segmentation knowledge acquired from the ROPS task, we transfer this knowledge to affordance detection by IRAS finetuning to enhance the model's generalization.
We also propose a sample unbalanced loss factor to address the learning strategies for affordance regions of different scales.
Specifically, during IRAS fine-tuning: we use the pretrained checkpoint \(W_\mathrm{f_{PB}}\) and \(W_\mathrm{f_{MD}}\) to initialize the modules \(f_\mathrm{PB}\) and \(f_\mathrm{AFD}\) in our framework 3D-ADLLM as shown in Fig.~\ref{model_architecture}. 
We then use the Lora method to fine-tune a pre-trained LLM for affordance segmentation.

\textbf{Training Objectives.}
The model is trained end-to-end using text generation loss \(\mathcal{L}_\mathrm{txt}\) and segmentation mask loss $\mathcal{L}_\mathrm{mask}$. 
Specifically,$\mathcal{L}_{txt}$ encourage the LLMs generate the response including the \texttt{<AFF>} token and forces the features to map to the same \texttt{<AFF>} placeholder. 
Then, $\mathcal{L} _ {mask}$ diversifies the \texttt{<AFF>} features to contain the information for affordance prediction and guides affordance mask generation.
The overall objective $\mathcal{L}$ is the weighted sum of these losses, determined by $\lambda_\mathrm{txt}$ and $\lambda_\mathrm{mask}$: $$\mathcal{L}=\lambda_\mathrm{txt}\mathcal{L}_\mathrm{txt}+\lambda_\mathrm{mask}\mathcal{L}_\mathrm{mask}.$$ Specifically, $\mathcal{L}_\mathrm{txt}$ is the auto-regressive cross-entropy loss for text generation, and $\mathcal{L}_\mathrm{mask}$ is the mask loss for high-quality segmentation. 
To compute $\mathcal{L}_\mathrm{mask}$, we use a combination of per-pixel BCE loss and DICE Loss, with weights $\lambda_\mathrm{bce}$ and $\lambda_\mathrm{dice}$. 
Given the ground-truth targets $\mathbf{y}_\mathrm{txt}$ and M obtained from dataset, these losses are formulated as: $$\mathcal{L}_\mathrm{txt}=\mathbf{CE}(\hat{\mathbf{y}}_\mathrm{txt},\mathbf{y}_\mathrm{txt}),$$ $$\mathcal{L}_\mathrm{mask}=\lambda_\mathrm{bce}\mathbf{BCE}(\hat{\mathbf{M}},\mathbf{M})+\lambda_\mathrm{dice}\mathbf{DICE}(\hat{\mathbf{M}},\mathbf{M}).$$

\textbf{Sample Unbalanced Loss Factor.}
Due to the varying scales of target affordance regions, our model 3D-ADLLM naturally challenges model's adaptiveness at different scales.
This variability will results in an imbalance in the difficulty of learning samples between different affordance types during the training process.
To mitigate the issue of sample imbalance across different affordance types during training, we apply weights to the mask losses for each class. 
The weighted loss is defined as:
${\mathcal{L}_\mathrm{mask} = \sum_{i=1}^{n}\omega_i \mathcal{L}_{\text{mask}}^i}$. 
The weight $\omega_i$ is calculated as:
$$\omega_i = \left( \frac{\max\{c_1, c_2, \ldots , c_m, c_0\}}{c_i} \right)^{1/4}$$
where \(c_i\) is the number of ground truth points for class \(i\), and \(c_0\) denotes background points.
\section{Experiment}
\subsection{Experiment Setting}
\textbf{Network Architecture.}
We use Phi-3.5-mini-instruct ($f_\mathrm{llm}$)~\citep{abdin2024phi} as our base LLM. For the point encoder ($f_\mathrm{pe}$), we adopt Point-BERT~\citep{yu2022pointbert}, pre-trained with ULIP-2~\citep{xue2024ulip2} in the ModelNet dataset~\citep{vishwanath2009modelnet}. The projector layer ($f_\mathrm{proj}$) between the point encoder $f_\mathrm{pe}$ and the LLM $f_\mathrm{llm}$ is a linear layer.
Additionally, we utilize the Point Transformer~\citep{zhao2021point_transformer}as the backbone for our point segmentation model ($f_\mathrm{PB}$).

\textbf{Datasets.} 
As is mentioned in Sec.~\ref{sec:two stage training strategy}, our training data is made up of two types of task data: (1) \textit{Referring Object Part Segmentation Dataset:} we build this dataset on PartNet~\citep{Mo_2019_CVPRPartnet}, which contains 573,585 part instances across 25,571 3D models and 24 object categories. 
For pre-training, we split it into single-part segmentation instances.
(2) \textit{Instruction Reasoning Affordance Segmentation Dataset:} we meticulously compile a question-point affordance dataset with 42119 paired samples from 3D AffordanceNet dataset~\citep{deng20213daffordancenet}, covering 23 classes and 36 affordance types. The Detailed data settings (full or partial view) and the visualized data analysis can be seen  in Appendix Sect.~\ref{apendix:data analysis}.

\textbf{Baseline Models.}
We compare our method with the following recent methods for zero-shot learning in 3D point clouds: 
ZSLPC~\citep{cheraghian2019ZSLPC}, TZSLPC~\citep{cheraghian2020TZSLPC}, 3DGenZ~\citep{michele20213DGenZ},
OpenAD~\citep{nguyen2023open}, IAGNet~\citep{yang2023IAGNet}, LASO~\citep{li2024laso} and ShapeLLM~\citep{qi2024shapellm}.
Detailed baseline model explanation for experiments can be found in Appendix Sect.~\ref{appendix:detailed baseline model explanation}.

\textbf{Evaluation metrics.}
We divide the IRAS dataset following the split of OpenAD and evaluate the close-set and open-set of IRAS.
According to \cite{nguyen2023open}, we use three metrics to evaluate the results over all \textbf{classes}:
$\mathrm{mIoU}^\mathrm{c}$ (mean IoU over all classes), $\mathrm{Acc}^\mathrm{c}$ (overall accuracy over all points), and $\mathrm{mAcc}^\mathrm{c}$ (mean accuracy over all classes). 
However, unlike OpenAD, which includes the "none" category in the calculation of metrics, we only compute the 36 affordance types, excluding "none," as it has little comparative significance.
For a comprehensive evaluation versus existing methods, we additionally assess each instance across the entire dataset.
The specific evaluation metrics over all \textbf{instances}: $\mathrm{mIoU}^\mathrm{i}$ (mean IoU over all instance data), $\mathrm{mAcc}^\mathrm{i}$ (mean accuracy of points over all instance data), $\mathrm{mPrec}^\mathrm{i}$ (mean precision of points over all instance data), $\mathrm{mRec}^\mathrm{i}$ (mean recall of points over all instance data), $\mathrm{mAP}_{50}^\mathrm{i}$ (mean average precision at 50\% intersection over union).

\subsection{Experiment Results}
\subsubsection{Comparison Results}
\textbf{3D-ADLLM vs. Other Models.}
Table~\ref{tab:main result} demonstrates that our 3D-ADLLM achieves superior performance across both full and partial view tasks, as well as on all three evaluation metrics.
Notably, 3D AffordanceLLM significantly outperforms the runner-up model (LASO) in terms of mIoU, with improvements of 8.02\% and 7.19\% on the full and partial view tasks, respectively.
Compared to OpenAD, which predicts regions based on a fixed set of affordance labels, our method utilizes long-context understanding and reasoning for segmentation.
In experiment results, our method surpasses OpenAD in terms of mIoU 16.9\% (full-view) and 15.96\% (partial-view) separately across 18 affordance types.
Additionally, for metrics over all instance, we surpass the sota model (LASO) 23.38\% (full-view) and 24.93\% (partial-view) in mAP$_{50}$.
The comparison results on close-set detection can be found in Appendix Sect.~\ref{appendix:close set results comparison}. 

\begin{table}
  \centering
  \vspace{-4pt}
  \caption{Main results of 3D-ADLLM on zero-short open vocabulary detection. 
  The result is calculated over all classes. 
  The overall results of all comparative methods, the best results are in bold.~$\ast$ The method of ShapeLLM is tested without finetuning.
  }
  \label{tab:main result}
  \vspace{6pt}
  \resizebox{\textwidth}{!}{
    \begin{tabular}{lrrrrrr}
    \toprule
    \multirow{2}{*}{Method} & \multicolumn{3}{|c|}{Full-view} & \multicolumn{3}{c}{Partial-view} \\
    & \multicolumn{1}{|c}{$\mathrm{mIoU}^\mathrm{c}$} & \multicolumn{1}{c}{$\mathrm{Acc}^\mathrm{c}$} & \multicolumn{1}{c|}{$\mathrm{mAcc}^\mathrm{c}$} & \multicolumn{1}{c}{$\mathrm{mIoU}^\mathrm{c}$} & \multicolumn{1}{c}{$\mathrm{Acc}^\mathrm{c}$} & \multicolumn{1}{c}{$\mathrm{mAcc}^\mathrm{c}$} \\
    \midrule
    % \rowcolor{gray!20} \multicolumn{1}{l|}{TZSLPC} 
    \multicolumn{1}{l|}{\gray{TZSLPC~\citep{cheraghian2020TZSLPC}}}
    &\multicolumn{1}{c}{\gray{3.86}} & \multicolumn{1}{c}{--} &\multicolumn{1}{c|}{\gray{10.37}} &\multicolumn{1}{c}{\gray{4.14}} & \multicolumn{1}{c}{--} &\multicolumn{1}{c}{\gray{8.49}}  \\
    
    % \rowcolor{gray!20}\multicolumn{1}{l|}{3DGenZ} 
    \multicolumn{1}{l|}{\gray{3DGenZ~\citep{michele20213DGenZ}}}
    &\multicolumn{1}{c}{\gray{6.46}} & \multicolumn{1}{c}{--} &\multicolumn{1}{c|}{\gray{18.33}} &\multicolumn{1}{c}{\gray{6.03}} & \multicolumn{1}{c}{--} &\multicolumn{1}{c}{\gray{15.86}}  \\
    
    % \rowcolor{gray!20}\multicolumn{1}{l|}{ZSLPC} 
    \multicolumn{1}{l|}{\gray{ZSLPC~\citep{cheraghian2019ZSLPC}}}
    &\multicolumn{1}{c}{\gray{9.97}} & \multicolumn{1}{c}{--} &\multicolumn{1}{c|}{\gray{18.70}} &\multicolumn{1}{c}{\gray{9.52}} & \multicolumn{1}{c}{--} &\multicolumn{1}{c}{\gray{17.16}}  \\
    \midrule
    
    \multicolumn{1}{l|}{ShapeLLM$\ast$~\citep{qi2024shapellm}}
    & \multicolumn{1}{c}{0.88} & \multicolumn{1}{c}{0.28} & \multicolumn{1}{c|}{0.99}     & \multicolumn{1}{c}{1.49} & \multicolumn{1}{c}{1.35} & \multicolumn{1}{c}{1.70}  \\  
    
    \multicolumn{1}{l|}{OpenAD-PointNet++~\citep{nguyen2023open}} 
    & \multicolumn{1}{c}{13.53} & \multicolumn{1}{c}{3.97} & \multicolumn{1}{c|}{16.40} 
    & \multicolumn{1}{c}{11.29} & \multicolumn{1}{c}{2.41} & \multicolumn{1}{c}{13.88} \\
    
    \multicolumn{1}{l|}{OpenAD-DGCNN~\citep{nguyen2023open}} 
    & \multicolumn{1}{c}{11.15} & \multicolumn{1}{c}{3.84} & \multicolumn{1}{c|}{13.86} 
    & \multicolumn{1}{c}{8.04} & \multicolumn{1}{c}{1.58} & \multicolumn{1}{c}{9.85} \\

    \multicolumn{1}{l|}{IAGNet~\citep{yang2023IAGNet}} 
    % & \multicolumn{1}{c}{12.58} & \multicolumn{1}{c}{17.30} & \multicolumn{1}{c|}{19.83} 
    & \multicolumn{1}{c}{16.16} & \multicolumn{1}{c}{19.07} & \multicolumn{1}{c|}{23.92} 
    & \multicolumn{1}{c}{14.36} & \multicolumn{1}{c}{16.90} & \multicolumn{1}{c}{21.73} \\
    
    \multicolumn{1}{l|}{LASO~\citep{li2024laso}} 
    & \multicolumn{1}{c}{22.41} & \multicolumn{1}{c}{15.90} & \multicolumn{1}{c|}{30.22} 
    & \multicolumn{1}{c}{20.06} & \multicolumn{1}{c}{8.80} & \multicolumn{1}{c}{26.84} \\

   \midrule
    % \multicolumn{1}{l|}{Ours-Gemma} &       &       & \multicolumn{1}{c|}{} &       &       &  \\
    \multicolumn{1}{l|}{Ours-Qwen} 
    & \multicolumn{1}{c}{24.43} & \multicolumn{1}{c}{23.90} & \multicolumn{1}{c|}{35.45} 
    & \multicolumn{1}{c}{26.25} & \multicolumn{1}{c}{\textbf{29.5}} & \multicolumn{1}{c}{\textbf{41.57}} \\
    
    \multicolumn{1}{l|}{Ours-Phi} 
    & \multicolumn{1}{c}{\textbf{30.43}} & \multicolumn{1}{c}{\textbf{29.36}} & \multicolumn{1}{c|}{\textbf{47.78}} 
    & \multicolumn{1}{c}{\textbf{27.25}} & \multicolumn{1}{c}{27.87} & \multicolumn{1}{c}{39.04} \\
    
    \bottomrule
    \end{tabular}%
    }
\end{table}%

\subsubsection{Out-of-distribution results}
\label{sec:out of distribution result}
The test in out-of-distribution (ood) datasets is essential to assess the generalization capability of the model.
Thus, we constructed a new test dataset consisting of approximately 559 entries by filtering out some combinations of affordance-object that already existed in our IRAS dataset from the AffordPose dataset~\citep{jian2023affordpose}.
Compared to existing datasets,
this new dataset includes different types of affordances as well as unique affordance-object pairs, such as \textit{(twist, faucet), (lever, faucet), (press, dispenser)}, etc. 
As is shown in Table~\ref{tab: zero short on AffordPose dataset}, our approach
achieved the best zero-shot performance on this ood data.
\begin{table}
\caption{Zero-shot Open-vocabulary detection results on over \textbf{all instances.}}
\label{tab: zero-shot task result}
\begin{center}
\renewcommand{\arraystretch}{1.1}
\begin{tabular}{c | l c c c c c}
\toprule
\rotatebox{90}{} & Method & $\mathrm{mIoU}^\mathrm{i}$  & $\mathrm{mAcc}^\mathrm{i}$ & $\mathrm{mPrec}^\mathrm{i}$ & $\mathrm{mRec}^\mathrm{i}$ & $\mathrm{mAP}_{50}^\mathrm{i}$\\
\midrule
\multirow{4}{*}{\rotatebox{90}{Full-view}} 
& OpenAD-PointNet++  & {3.46} & {\textbf{74.59}} & {11.84} & {5.84} & {0.02} \\
& OpenAD-DGCNN       & {3.79} & {74.42} & {11.13} & {6.67} & {0.04} \\
& LASO               & {20.47} & {71.47} & {37.95} & {34.93} & {2.42} \\
& 3D-ADLLM (ours)    & \textbf{30.28} & 70.66 & \textbf{40.89} & \textbf{55.93} & \textbf{27.80} \\
\midrule
\multirow{4}{*}{\rotatebox{90}{Partial-view}} 
& OpenAD-PointNet++ & {2.17} & {71.97} & {5.64} & {3.74} & {0.02} \\
& OpenAD-DGCNN      & {2.08} & {72.00} & {6.65} & {3.40} & {0.02} \\
& LASO              & {11.46} & {\textbf{72.14}} & {32.70} & {16.49} & {0.70} \\
& 3D-ADLLM (ours)   & \textbf{28.72} & 68.28 & \textbf{41.71} & \textbf{47.73} & \textbf{25.63} \\
\bottomrule
\end{tabular}
\end{center}
\end{table}
\begin{table}
\caption{Zero-shot Open-vocabulary detection results on AffordPose data over \textbf{all instances.}}
\label{tab: zero short on AffordPose dataset}
\begin{center}
\begin{tabular}{lllcccc}
\toprule
Method & $\mathrm{mIoU}^\mathrm{i}$  & $\mathrm{mAcc}^\mathrm{i}$ & $\mathrm{mPrec}^\mathrm{i}$ & $\mathrm{mRec}^\mathrm{i}$ & $\mathrm{mAP}_{50}^\mathrm{i}$\\
\midrule
OpenAD-PointNet++ &7.61  & 65.13 &22.47 &13.01&0.37\\
OpenAD-DGCNN &8.02  &66.76  & 15.83 & 13.52 &0.39\\
LASO &34.49  &\textbf{77.12 } & \textbf{56.04} & 37.88 &8.40\\
3D-ADLLM (ours) &\textbf{36.33} &74.79 &55.46  &\textbf{46.80} &\textbf{36.33}\\
\bottomrule
\end{tabular}
\end{center}
\end{table}

\subsection{Ablation Study}
\textbf{Effects of Different Components.}
To investigate the effectiveness of each component in 3D-ADLLM, we conduct experiments with different variants of 3D-ADLLM. In particular, we compare 2 different implementations: (1) w/o PC removes the
pre-trained weights \(f_\mathrm{PB}\) and \(f_\mathrm{AFD}\), directly training our 3D-ADLLM; (2) w/o UL removes the
sample unbalanced factor.
As is shown in Table~\ref{tab:main ablation},
the performance of 3D-ADLLM drops significantly without either of these components.
Notably, the most substantial performance degradation with about 6\% occurs in mIoU when the PC module is removed. UL is also critical for our framework. Once it is removed, the performance, there is a noticeable reduction in the model’s performance.

\textbf{Effects of Different Backbones.}
As shown in Table \ref{tab:main result}, we experimented with different LLM backbones to evaluate the effectiveness of our framework.
Specifically, we chose Phi-3.5-mini-instruct~\citep{abdin2024phi} and Qwen2-1.5B~\citep{yang2024qwen2} as the LLM backbone.
\begin{wraptable}{r}{0.4\textwidth}
\vspace{-2.5mm}
    \caption{The efforts with different point encoder $f_\mathrm{pe}$ in 3D-ADLLM.(Full-View)}
    \label{tab:study of point encoder}
    \resizebox{1\linewidth}{!}{
    \begin{tabular}{lccc}
    \toprule
    $f_\mathrm{pe}$ & $\mathrm{mIoU}^\mathrm{c}$  & $\mathrm{Acc}^\mathrm{c}$ & $\mathrm{mAcc}^\mathrm{c}$ \\
    \midrule
    ULIP2 & \textbf{30.43} &\textbf{29.36} &47.78 \\
    Uni3D & 30.26  &26.21  &\textbf{48.16 } \\
    \bottomrule
    \end{tabular}
    }
    \vspace{-3mm}
\end{wraptable}
In terms of experimental results, Phi outperforms Qwen in the full-view setting. However, in the partial-view setting, the performance of Phi shows no significant difference compared to Qwen. 
Based on these findings, 3D-ADLLM adopts Phi as the default LLM backbone. 
In addition to testing different LLM backbones, we also explored different point encoders. Table~\ref{tab:study of point encoder} summarizes the performance of ULIP2~\citep{xue2024ulip2} and Uni3D~\citep{zhou2023uni3d} as point encoders, while ULIP2 obtained slightly better mean accuracy.

\textbf{Effects of Different Learning Objectives.}
We define the Affordance Region Ratio ($\mathrm{Arr}$) as ${p}_\mathrm{aff}/{{p}_\mathrm{cloud}}$, representing the proportion of affordance regions relative to the point cloud. 
In the IRAS task, the average $\mathrm{Arr}$ is approximately 18\%. 
However, for specific categories like "pull" and "listen," it is around 5\%, while for "wear" it reaches about 40\%. 
Variations in $\mathrm{Arr}$ across different predictions lead to class imbalances. 
Dice Loss, a segmentation loss function, measures the similarity between predictions and ground truth. 
Unlike Binary Cross-Entropy Loss (BCE), which focuses on pixel-level differences, Dice Loss emphasizes global region similarity, making it more effective for handling data imbalance. 
As shown in Table~\ref{tab:loss weight}, the model utilizing Dice Loss achieves superior mIoU metrics in both seen and unseen settings. Table~\ref{tab:loss weight} demonstrates that while the exclusive application of Dice Loss yields a marginal improvement on unseen data, it does not perform as well on seen data when compared to the combined usage of Dice Loss and BCE Loss.

\begin{table}[htbp]
\begin{center}
\begin{minipage}{0.495\textwidth}
\centering
\caption{The comparison results regarding different settings of loss.(full-view)}
\vspace{2pt}
\resizebox{\textwidth}{!}{
\begin{tabular}{lcc}
\toprule
& Openset-$\mathrm{mIoU}^\mathrm{c}$  & Closeset-$\mathrm{mIoU}^\mathrm{c}$ \\
\midrule
DICE \& BCE & 30.43 & \textbf{42.35} \\
DICE & \textbf{31.00} & 38.65 \\
BCE & 15.99 & 31.14 \\
\bottomrule
\end{tabular}
}
\label{tab:loss weight}
\end{minipage}
\hfill
\begin{minipage}{0.46\textwidth}
\centering
\caption{Results of 3D-ADLLM variants with removing different components.(full-view)}
 \vspace{1pt}
% \caption{Ablation study with different components.}
\resizebox{\textwidth}{!}{
\begin{tabular}{lllccccc}
\toprule
Model & $\mathrm{mIoU}^\mathrm{c}$  & $\mathrm{Acc}^\mathrm{c}$ & $\mathrm{mAcc}^\mathrm{c}$ \\
\midrule
3D-ADLLM & \textbf{30.43} & \textbf{29.36} & \textbf{47.78} \\
3D-ADLLM$_{\mathrm{~w/o~PC}}$ & 24.82 & 20.54 & 36.73 \\
3D-ADLLM$_{\mathrm{~w/o~UL}}$ & 25.35 & 26.84 & 40.69 \\
\bottomrule
\end{tabular}
}
\label{tab:main ablation}
\end{minipage}
\end{center}
\end{table}
% \vspace{-3mm}
\subsection{Qualitative Results}
As shown in Fig.~\ref{fig:visualization comparision}, our model demonstrates the capacity to accurately comprehend object affor-
dance given the complex reasoning instruction.
It is noteworthy that even when dealing with small affordance components, such as the  switch of faucet, our model still exhibits decent ability. 
Moreover, our 3D-ADLLM surpasses other models by employing a multi-stage training strategy that facilitates knowledge transfer and extraction of world knowledge from LLMs.
For example, when identifying areas on a chair that can take a seat Fig.~\ref{fig:visualization comparision}~(e) or areas that can wrap around a cup Fig.~\ref{fig:visualization comparision}~(g), our model significantly outperforms other models.
\begin{figure}[htbp]
    \centering 
    % \vspace{-5pt}
    \includegraphics[width=1.0\linewidth]{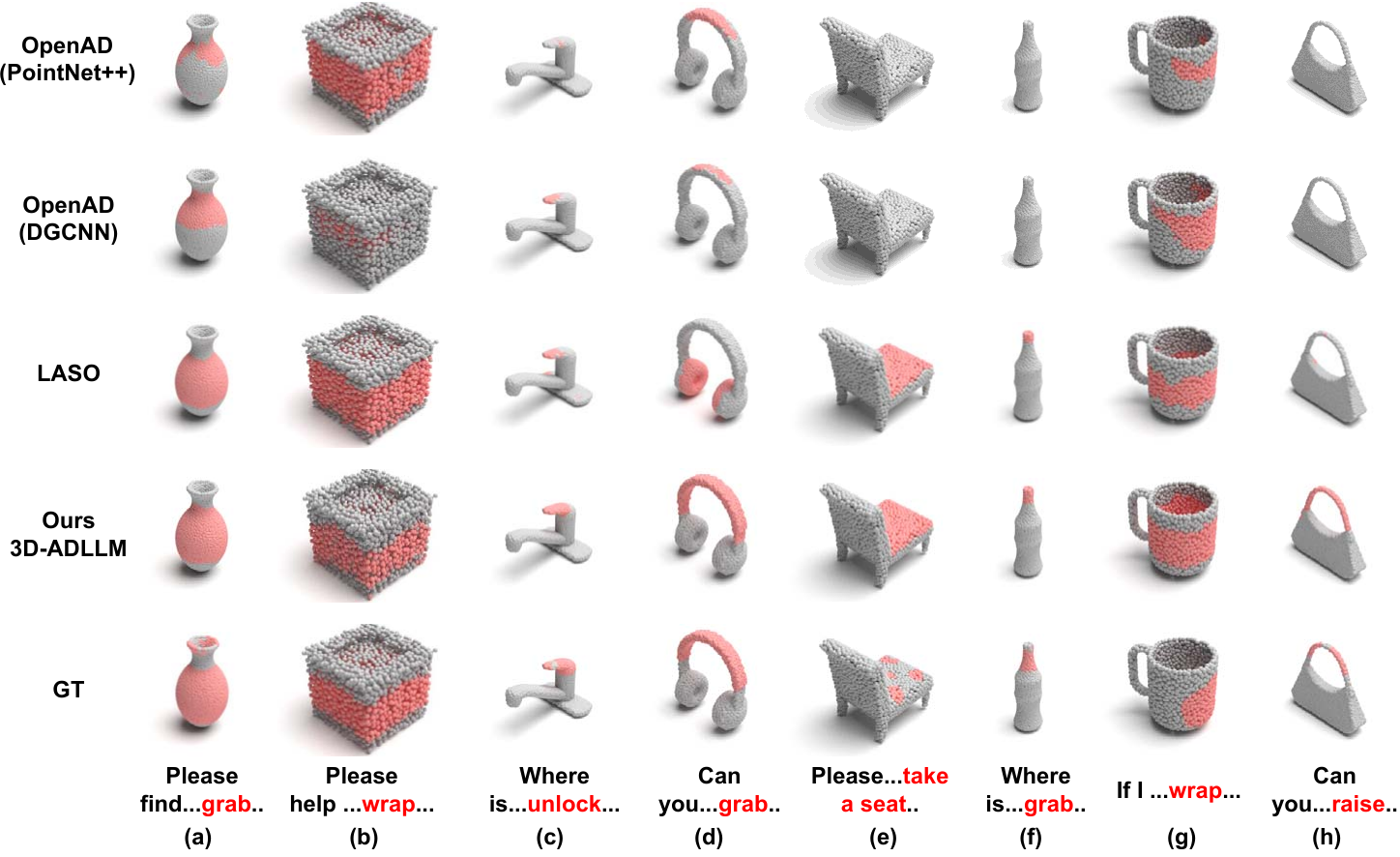}
    \vspace{-8pt}
    \caption{The visualization results of our 3D-ADLLM compared with others.}
    \label{fig:visualization comparision}
    \vspace{-10pt}
\end{figure}
% \vspace{-3mm}
\section{Conclusion}
In this work, we reformulate the traditional affordance detection paradigm into \textit{Instruction Reasoning Affordance Segmentation} (IRAS) task, enabling open-world affordance detection.
Then, we propose the multi-stage learning strategy with a novel defined Referring Object Part Segmentation (ROPS) task to extract general segmentation knowledge to affordance detection.
Finally, we accordingly proposed the 3D-AffordanceLLM (3D-ADLLM), firstly injecting LLM into 3D affordance perception, a framework designed for query reasoning affordance segmentation in 3D open scenarios. 
Experimental results demonstrate the effectiveness of 3D-ADLLM, we hope our work can shed new light on the direction of affordance detection in open-world scene in the future.
% \normalem
\section{Acknowledgement}
We would like to thank the reviewers for their constructive comments. This study is supported by Shenzhen College Stability Support Plan (Grant No. GXWD20220817144428005), National Natural Science Foundation of China (Grant No. 62406092), National Natural Science Foundation of China (Grant No. U24B20175), Shenzhen Science and Technology Program (Grant No. KJZD20240903100017022), Research on Efficient Exploration and Self-Evolution of APP Agents \& Embodied Intelligent Cerebellum Control Model and Collaborative Feedback Training Project (Grant No. TC20240403047).

\bibliography{iclr2025_conference}

\begin{thebibliography}{48}
\providecommand{\natexlab}[1]{#1}
\providecommand{\url}[1]{\texttt{#1}}
\expandafter\ifx\csname urlstyle\endcsname\relax
  \providecommand{\doi}[1]{doi: #1}\else
  \providecommand{\doi}{doi: \begingroup \urlstyle{rm}\Url}\fi

\bibitem[Abdin et~al.(2024)Abdin, Jacobs, Awan, Aneja, Awadallah, Awadalla, Bach, Bahree, Bakhtiari, Behl, et~al.]{abdin2024phi}
Marah Abdin, Sam~Ade Jacobs, Ammar~Ahmad Awan, Jyoti Aneja, Ahmed Awadallah, Hany Awadalla, Nguyen Bach, Amit Bahree, Arash Bakhtiari, Harkirat Behl, et~al.
\newblock Phi-3 technical report: A highly capable language model locally on your phone.
\newblock \emph{arXiv preprint arXiv:2404.14219}, 2024.

\bibitem[Chen et~al.(2023)Chen, Gao, Lin, and Shou]{chen2023affordance}
Joya Chen, Difei Gao, Kevin~Qinghong Lin, and Mike~Zheng Shou.
\newblock Affordance grounding from demonstration video to target image.
\newblock In \emph{Proceedings of the IEEE/CVF Conference on Computer Vision and Pattern Recognition}, pp.\  6799--6808, 2023.

\bibitem[Chen et~al.(2024)Chen, Chen, Zhang, Li, Yu, Fei, Zhu, Fan, and Chen]{chen2024ll3da}
Sijin Chen, Xin Chen, Chi Zhang, Mingsheng Li, Gang Yu, Hao Fei, Hongyuan Zhu, Jiayuan Fan, and Tao Chen.
\newblock Ll3da: Visual interactive instruction tuning for omni-3d understanding reasoning and planning.
\newblock In \emph{Proceedings of the IEEE/CVF Conference on Computer Vision and Pattern Recognition}, pp.\  26428--26438, 2024.

\bibitem[Cheraghian et~al.(2019)Cheraghian, Rahman, and Petersson]{cheraghian2019ZSLPC}
Ali Cheraghian, Shafin Rahman, and Lars Petersson.
\newblock Zero-shot learning of 3d point cloud objects.
\newblock In \emph{2019 16th International Conference on Machine Vision Applications (MVA)}, pp.\  1--6. IEEE, 2019.

\bibitem[Cheraghian et~al.(2020)Cheraghian, Rahman, Campbell, and Petersson]{cheraghian2020TZSLPC}
Ali Cheraghian, Shafin Rahman, Dylan Campbell, and Lars Petersson.
\newblock Transductive zero-shot learning for 3d point cloud classification.
\newblock In \emph{Proceedings of the IEEE/CVF Winter Conference on Applications of Computer Vision}, pp.\  923--933, 2020.

\bibitem[Deitke et~al.(2023)Deitke, Schwenk, Salvador, Weihs, Michel, VanderBilt, Schmidt, Ehsani, Kembhavi, and Farhadi]{deitke2023objaverse}
Matt Deitke, Dustin Schwenk, Jordi Salvador, Luca Weihs, Oscar Michel, Eli VanderBilt, Ludwig Schmidt, Kiana Ehsani, Aniruddha Kembhavi, and Ali Farhadi.
\newblock Objaverse: A universe of annotated 3d objects.
\newblock In \emph{Proceedings of the IEEE/CVF Conference on Computer Vision and Pattern Recognition}, pp.\  13142--13153, 2023.

\bibitem[Deng et~al.(2021)Deng, Xu, Wu, Chen, and Jia]{deng20213daffordancenet}
Shengheng Deng, Xun Xu, Chaozheng Wu, Ke~Chen, and Kui Jia.
\newblock 3d affordancenet: A benchmark for visual object affordance understanding.
\newblock In \emph{proceedings of the IEEE/CVF Conference on Computer Vision and Pattern Recognition}, pp.\  1778--1787, 2021.

\bibitem[Do et~al.(2018)Do, Nguyen, and Reid]{do2018affordancenet}
Thanh-Toan Do, Anh Nguyen, and Ian Reid.
\newblock Affordancenet: An end-to-end deep learning approach for object affordance detection.
\newblock In \emph{2018 IEEE international conference on robotics and automation (ICRA)}, pp.\  5882--5889. IEEE, 2018.

\bibitem[Geng et~al.(2023)Geng, An, Geng, Chen, Yang, and Dong]{geng2023rlafford}
Yiran Geng, Boshi An, Haoran Geng, Yuanpei Chen, Yaodong Yang, and Hao Dong.
\newblock Rlafford: End-to-end affordance learning for robotic manipulation.
\newblock In \emph{2023 IEEE International Conference on Robotics and Automation (ICRA)}, pp.\  5880--5886. IEEE, 2023.

\bibitem[Gibson(1966)]{Gibson1966-GIBTSC-5}
James~Jerome Gibson.
\newblock \emph{The Senses Considered as Perceptual Systems}.
\newblock Houghton Mifflin, Boston, USA, 1966.

\bibitem[Hong et~al.(2023{\natexlab{a}})Hong, Zhen, Chen, Zheng, Du, Chen, and Gan]{hong20233d}
Yining Hong, Haoyu Zhen, Peihao Chen, Shuhong Zheng, Yilun Du, Zhenfang Chen, and Chuang Gan.
\newblock 3d-llm: Injecting the 3d world into large language models.
\newblock \emph{Advances in Neural Information Processing Systems}, 36:\penalty0 20482--20494, 2023{\natexlab{a}}.

\bibitem[Hong et~al.(2023{\natexlab{b}})Hong, Zhen, Chen, Zheng, Du, Chen, and Gan]{hong20233dllm}
Yining Hong, Haoyu Zhen, Peihao Chen, Shuhong Zheng, Yilun Du, Zhenfang Chen, and Chuang Gan.
\newblock 3d-llm: Injecting the 3d world into large language models.
\newblock \emph{Advances in Neural Information Processing Systems}, 36:\penalty0 20482--20494, 2023{\natexlab{b}}.

\bibitem[Hou et~al.(2021)Hou, Yu, Qiao, Peng, and Tao]{hou2021affordance}
Zhi Hou, Baosheng Yu, Yu~Qiao, Xiaojiang Peng, and Dacheng Tao.
\newblock Affordance transfer learning for human-object interaction detection.
\newblock In \emph{Proceedings of the IEEE/CVF Conference on Computer Vision and Pattern Recognition}, pp.\  495--504, 2021.

\bibitem[Huang et~al.(2023)Huang, Wang, Huang, Liu, Cheng, Zhao, Jin, and Zhao]{huang2023chat3dv2}
Haifeng Huang, Zehan Wang, Rongjie Huang, Luping Liu, Xize Cheng, Yang Zhao, Tao Jin, and Zhou Zhao.
\newblock Chat-3d v2: Bridging 3d scene and large language models with object identifiers.
\newblock \emph{arXiv preprint arXiv:2312.08168}, 2023.

\bibitem[Jian et~al.(2023)Jian, Liu, Li, Hu, and Liu]{jian2023affordpose}
Juntao Jian, Xiuping Liu, Manyi Li, Ruizhen Hu, and Jian Liu.
\newblock Affordpose: A large-scale dataset of hand-object interactions with affordance-driven hand pose.
\newblock In \emph{Proceedings of the IEEE/CVF International Conference on Computer Vision}, pp.\  14713--14724, 2023.

\bibitem[Krizhevsky et~al.(2012)Krizhevsky, Sutskever, and Hinton]{krizhevsky2012imagenet}
Alex Krizhevsky, Ilya Sutskever, and Geoffrey~E Hinton.
\newblock Imagenet classification with deep convolutional neural networks.
\newblock \emph{Advances in neural information processing systems}, 25, 2012.

\bibitem[Lai et~al.(2024)Lai, Tian, Chen, Li, Yuan, Liu, and Jia]{lai2024lisa}
Xin Lai, Zhuotao Tian, Yukang Chen, Yanwei Li, Yuhui Yuan, Shu Liu, and Jiaya Jia.
\newblock Lisa: Reasoning segmentation via large language model.
\newblock In \emph{Proceedings of the IEEE/CVF Conference on Computer Vision and Pattern Recognition}, pp.\  9579--9589, 2024.

\bibitem[Li et~al.(2023)Li, Jampani, Sun, and Sevilla-Lara]{li2023locate}
Gen Li, Varun Jampani, Deqing Sun, and Laura Sevilla-Lara.
\newblock Locate: Localize and transfer object parts for weakly supervised affordance grounding.
\newblock In \emph{Proceedings of the IEEE/CVF Conference on Computer Vision and Pattern Recognition}, pp.\  10922--10931, 2023.

\bibitem[Li et~al.(2024)Li, Zhao, Xiao, Feng, Wang, and Chua]{li2024laso}
Yicong Li, Na~Zhao, Junbin Xiao, Chun Feng, Xiang Wang, and Tat-seng Chua.
\newblock Laso: Language-guided affordance segmentation on 3d object.
\newblock In \emph{Proceedings of the IEEE/CVF Conference on Computer Vision and Pattern Recognition}, pp.\  14251--14260, 2024.

\bibitem[Liu et~al.(2019)Liu, Sun, Li, Hu, and Wang]{liu2019deep}
Weiping Liu, Jia Sun, Wanyi Li, Ting Hu, and Peng Wang.
\newblock Deep learning on point clouds and its application: A survey.
\newblock \emph{Sensors}, 19\penalty0 (19):\penalty0 4188, 2019.

\bibitem[Lu et~al.(2022)Lu, Zhai, Luo, Kang, and Cao]{lu2022phrase}
Liangsheng Lu, Wei Zhai, Hongchen Luo, Yu~Kang, and Yang Cao.
\newblock Phrase-based affordance detection via cyclic bilateral interaction.
\newblock \emph{IEEE Transactions on Artificial Intelligence}, 4\penalty0 (5):\penalty0 1186--1198, 2022.

\bibitem[Luo et~al.(2022)Luo, Zhai, Zhang, Cao, and Tao]{luo2022learning}
Hongchen Luo, Wei Zhai, Jing Zhang, Yang Cao, and Dacheng Tao.
\newblock Learning affordance grounding from exocentric images.
\newblock In \emph{Proceedings of the IEEE/CVF Conference on Computer Vision and Pattern Recognition}, pp.\  2252--2261, 2022.

\bibitem[Matheson et~al.(2019)Matheson, Minto, Zampieri, Faccio, and Rosati]{matheson2019human}
Eloise Matheson, Riccardo Minto, Emanuele~GG Zampieri, Maurizio Faccio, and Giulio Rosati.
\newblock Human--robot collaboration in manufacturing applications: A review.
\newblock \emph{Robotics}, 8\penalty0 (4):\penalty0 100, 2019.

\bibitem[Mi et~al.(2020)Mi, Liang, Katsakis, Tang, Li, Zhang, and Zhang]{mi2020intention}
Jinpeng Mi, Hongzhuo Liang, Nikolaos Katsakis, Song Tang, Qingdu Li, Changshui Zhang, and Jianwei Zhang.
\newblock Intention-related natural language grounding via object affordance detection and intention semantic extraction.
\newblock \emph{Frontiers in Neurorobotics}, 14:\penalty0 26, 2020.

\bibitem[Michele et~al.(2021)Michele, Boulch, Puy, Bucher, and Marlet]{michele20213DGenZ}
Bj{\"o}rn Michele, Alexandre Boulch, Gilles Puy, Maxime Bucher, and Renaud Marlet.
\newblock Generative zero-shot learning for semantic segmentation of 3d point clouds.
\newblock In \emph{2021 International Conference on 3D Vision (3DV)}, pp.\  992--1002. IEEE, 2021.

\bibitem[Min et~al.(2016)Min, Luo, Zhu, Bi, et~al.]{min2016affordance}
Huaqing Min, Ronghua Luo, Jinhui Zhu, Sheng Bi, et~al.
\newblock Affordance research in developmental robotics: A survey.
\newblock \emph{IEEE Transactions on Cognitive and Developmental Systems}, 8\penalty0 (4):\penalty0 237--255, 2016.

\bibitem[Mo et~al.(2019)Mo, Zhu, Chang, Yi, Tripathi, Guibas, and Su]{Mo_2019_CVPRPartnet}
Kaichun Mo, Shilin Zhu, Angel~X. Chang, Li~Yi, Subarna Tripathi, Leonidas~J. Guibas, and Hao Su.
\newblock {PartNet}: A large-scale benchmark for fine-grained and hierarchical part-level {3D} object understanding.
\newblock In \emph{The IEEE Conference on Computer Vision and Pattern Recognition (CVPR)}, June 2019.

\bibitem[Mo et~al.(2022)Mo, Qin, Xiang, Su, and Guibas]{mo2022o2o}
Kaichun Mo, Yuzhe Qin, Fanbo Xiang, Hao Su, and Leonidas Guibas.
\newblock O2o-afford: Annotation-free large-scale object-object affordance learning.
\newblock In \emph{Conference on robot learning}, pp.\  1666--1677. PMLR, 2022.

\bibitem[Moldovan et~al.(2012)Moldovan, Moreno, Van~Otterlo, Santos-Victor, and De~Raedt]{moldovan2012learning}
Bogdan Moldovan, Plinio Moreno, Martijn Van~Otterlo, Jos{\'e} Santos-Victor, and Luc De~Raedt.
\newblock Learning relational affordance models for robots in multi-object manipulation tasks.
\newblock In \emph{2012 ieee International Conference on Robotics and Automation}, pp.\  4373--4378. IEEE, 2012.

\bibitem[Nagarajan et~al.(2019)Nagarajan, Feichtenhofer, and Grauman]{nagarajan2019grounded}
Tushar Nagarajan, Christoph Feichtenhofer, and Kristen Grauman.
\newblock Grounded human-object interaction hotspots from video.
\newblock In \emph{Proceedings of the IEEE/CVF International Conference on Computer Vision}, pp.\  8688--8697, 2019.

\bibitem[Nguyen et~al.(2016)Nguyen, Kanoulas, Caldwell, and Tsagarakis]{nguyen2016detecting}
Anh Nguyen, Dimitrios Kanoulas, Darwin~G Caldwell, and Nikos~G Tsagarakis.
\newblock Detecting object affordances with convolutional neural networks.
\newblock In \emph{2016 IEEE/RSJ International Conference on Intelligent Robots and Systems (IROS)}, pp.\  2765--2770. IEEE, 2016.

\bibitem[Nguyen et~al.(2023)Nguyen, Vu, Vuong, Nguyen, Vo, Le, and Nguyen]{nguyen2023open}
Toan Nguyen, Minh~Nhat Vu, An~Vuong, Dzung Nguyen, Thieu Vo, Ngan Le, and Anh Nguyen.
\newblock Open-vocabulary affordance detection in 3d point clouds.
\newblock In \emph{2023 IEEE/RSJ International Conference on Intelligent Robots and Systems (IROS)}, pp.\  5692--5698. IEEE, 2023.

\bibitem[Pacheco-Ortega \& Mayol-Cuervas(2022)Pacheco-Ortega and Mayol-Cuervas]{pacheco2022one}
Abel Pacheco-Ortega and Walterio Mayol-Cuervas.
\newblock One-shot learning for human affordance detection.
\newblock In \emph{European Conference on Computer Vision}, pp.\  758--766. Springer, 2022.

\bibitem[Qi et~al.(2017)Qi, Yi, Su, and Guibas]{qi2017pointnet++}
Charles~Ruizhongtai Qi, Li~Yi, Hao Su, and Leonidas~J Guibas.
\newblock Pointnet++: Deep hierarchical feature learning on point sets in a metric space.
\newblock \emph{Advances in neural information processing systems}, 30, 2017.

\bibitem[Qi et~al.(2024)Qi, Dong, Zhang, Geng, Han, Ge, Yi, and Ma]{qi2024shapellm}
Zekun Qi, Runpei Dong, Shaochen Zhang, Haoran Geng, Chunrui Han, Zheng Ge, Li~Yi, and Kaisheng Ma.
\newblock Shapellm: Universal 3d object understanding for embodied interaction.
\newblock \emph{arXiv preprint arXiv:2402.17766}, 2024.

\bibitem[Radford et~al.(2021)Radford, Kim, Hallacy, Ramesh, Goh, Agarwal, Sastry, Askell, Mishkin, Clark, et~al.]{radford2021cliplearning}
Alec Radford, Jong~Wook Kim, Chris Hallacy, Aditya Ramesh, Gabriel Goh, Sandhini Agarwal, Girish Sastry, Amanda Askell, Pamela Mishkin, Jack Clark, et~al.
\newblock Learning transferable visual models from natural language supervision.
\newblock In \emph{International Conference on Machine Learning}, pp.\  8748--8763. PMLR, 2021.

\bibitem[Roy \& Fernando(2021)Roy and Fernando]{roy2021action}
Debaditya Roy and Basura Fernando.
\newblock Action anticipation using pairwise human-object interactions and transformers.
\newblock \emph{IEEE Transactions on Image Processing}, 30:\penalty0 8116--8129, 2021.

\bibitem[Vaswani et~al.(2017)Vaswani, Shazeer, Parmar, Uszkoreit, Jones, Gomez, Kaiser, and Polosukhin]{AshishVaswani16}
Ashish Vaswani, Noam Shazeer, Niki Parmar, Jakob Uszkoreit, Llion Jones, Aidan~N Gomez, \L~ukasz Kaiser, and Illia Polosukhin.
\newblock Attention is all you need.
\newblock \emph{NeurIPS}, 30, 2017.

\bibitem[Vishwanath et~al.(2009)Vishwanath, Gupta, Vahdat, and Yocum]{vishwanath2009modelnet}
Kashi~Venkatesh Vishwanath, Diwaker Gupta, Amin Vahdat, and Ken Yocum.
\newblock Modelnet: Towards a datacenter emulation environment.
\newblock In \emph{2009 IEEE Ninth International Conference on Peer-to-Peer Computing}, pp.\  81--82. IEEE, 2009.

\bibitem[Wang et~al.(2019)Wang, Sun, Liu, Sarma, Bronstein, and Solomon]{wang2019dynamic}
Yue Wang, Yongbin Sun, Ziwei Liu, Sanjay~E Sarma, Michael~M Bronstein, and Justin~M Solomon.
\newblock Dynamic graph cnn for learning on point clouds.
\newblock \emph{ACM Transactions on Graphics (tog)}, 38\penalty0 (5):\penalty0 1--12, 2019.

\bibitem[Wang et~al.(2023)Wang, Huang, Zhao, Zhang, and Zhao]{wang2023chat3d}
Zehan Wang, Haifeng Huang, Yang Zhao, Ziang Zhang, and Zhou Zhao.
\newblock Chat-3d: Data-efficiently tuning large language model for universal dialogue of 3d scenes.
\newblock \emph{arXiv preprint arXiv:2308.08769}, 2023.

\bibitem[Xue et~al.(2023)Xue, Gao, Xing, Mart{\'\i}n-Mart{\'\i}n, Wu, Xiong, Xu, Niebles, and Savarese]{xue2023ulip}
Le~Xue, Mingfei Gao, Chen Xing, Roberto Mart{\'\i}n-Mart{\'\i}n, Jiajun Wu, Caiming Xiong, Ran Xu, Juan~Carlos Niebles, and Silvio Savarese.
\newblock Ulip: Learning a unified representation of language, images, and point clouds for 3d understanding.
\newblock In \emph{Proceedings of the IEEE/CVF Conference on Computer Vision and Pattern Recognition}, pp.\  1179--1189, 2023.

\bibitem[Xue et~al.(2024)Xue, Yu, Zhang, Panagopoulou, Li, Mart{\'\i}n-Mart{\'\i}n, Wu, Xiong, Xu, Niebles, et~al.]{xue2024ulip2}
Le~Xue, Ning Yu, Shu Zhang, Artemis Panagopoulou, Junnan Li, Roberto Mart{\'\i}n-Mart{\'\i}n, Jiajun Wu, Caiming Xiong, Ran Xu, Juan~Carlos Niebles, et~al.
\newblock Ulip-2: Towards scalable multimodal pre-training for 3d understanding.
\newblock In \emph{Proceedings of the IEEE/CVF Conference on Computer Vision and Pattern Recognition}, pp.\  27091--27101, 2024.

\bibitem[Yang et~al.(2024)Yang, Yang, Hui, Zheng, Yu, Zhou, Li, Li, Liu, Huang, et~al.]{yang2024qwen2}
An~Yang, Baosong Yang, Binyuan Hui, Bo~Zheng, Bowen Yu, Chang Zhou, Chengpeng Li, Chengyuan Li, Dayiheng Liu, Fei Huang, et~al.
\newblock Qwen2 technical report.
\newblock \emph{arXiv preprint arXiv:2407.10671}, 2024.

\bibitem[Yang et~al.(2023)Yang, Zhai, Luo, Cao, Luo, and Zha]{yang2023IAGNet}
Yuhang Yang, Wei Zhai, Hongchen Luo, Yang Cao, Jiebo Luo, and Zheng-Jun Zha.
\newblock Grounding 3d object affordance from 2d interactions in images.
\newblock In \emph{Proceedings of the IEEE/CVF International Conference on Computer Vision}, pp.\  10905--10915, 2023.

\bibitem[Yu et~al.(2022)Yu, Tang, Rao, Huang, Zhou, and Lu]{yu2022pointbert}
Xumin Yu, Lulu Tang, Yongming Rao, Tiejun Huang, Jie Zhou, and Jiwen Lu.
\newblock Point-bert: Pre-training 3d point cloud transformers with masked point modeling.
\newblock In \emph{Proceedings of the IEEE/CVF Conference on Computer Vision and Pattern Recognition}, pp.\  19313--19322, 2022.

\bibitem[Zhao et~al.(2021)Zhao, Jiang, Jia, Torr, and Koltun]{zhao2021point_transformer}
Hengshuang Zhao, Li~Jiang, Jiaya Jia, Philip~HS Torr, and Vladlen Koltun.
\newblock Point transformer.
\newblock In \emph{Proceedings of the IEEE/CVF Conference on Computer Vision and Pattern Recognition}, pp.\  16259--16268, 2021.

\bibitem[Zhou et~al.(2023)Zhou, Wang, Ma, Liu, Huang, and Wang]{zhou2023uni3d}
Junsheng Zhou, Jinsheng Wang, Baorui Ma, Yu-Shen Liu, Tiejun Huang, and Xinlong Wang.
\newblock Uni3d: Exploring unified 3d representation at scale.
\newblock \emph{arXiv preprint arXiv:2310.06773}, 2023.

\end{thebibliography}
\bibliographystyle{iclr2025_conference}

\appendix
\section{Appendix}
\subsection{Baseline Models Details}
\label{appendix:detailed baseline model explanation}
We compare our method with the following recent methods for zero-shot learning in 3D point clouds: ZSLPC~\citep{cheraghian2019ZSLPC}, TZSLPC~\citep{cheraghian2020TZSLPC}, and 3DGenZ~\citep{michele20213DGenZ}. 
For these baselines, we change their original text encoders with CLIP and retain the same settings in OpenAD~\citep{nguyen2023open}. Furthermore, we incorporate two affordance detection works (IAGNet~\citep{yang2023IAGNet} and LASO~\citep{li2024laso}) to provide a more comprehensive comparison of our approach.
For IAGNet~\citep{yang2023IAGNet}, an affordance detection method that utilizes paired image-point cloud data as input. 
To tailor IAGNet~\citep{yang2023IAGNet} to our requirements, we seamlessly integrate a language model in place of its original image backbone, while maintaining the rest of its architecture unchanged. 
ShapeLLM-7B~\citep{qi2024shapellm} is a large-scale point cloud model that accepts point cloud and natural language inputs and possesses grounding capabilities. 
Consequently, we leverage its grounding abilities to perform zero-shot detection and calculate masks for comparison.

\subsection{Comparison results on close set}
In this work, we primarily focus on enhancing affordance detection capabilities in open-world scene. However, our model still performs well in closed-set affordance detection tasks. As is shown in Fig.~\ref{appendix:close set results comparison over all classes}, Fig.~\ref{appendix: close-set task result over all instances}, our 3D-ADLLM achieves optimal performance on nearly all metrics in both the over-all-classes and over-all-instances settings.

\label{appendix:close set results comparison}
\begin{table}[htbp]
  \centering
  \caption{Main results of 3D-ADLLM compared with other methods on close-set detection over all class.}
  \label{appendix:close set results comparison over all classes}
  \vspace{8pt}
  \resizebox{\textwidth}{!}{
    \begin{tabular}{lrrrrrr}
    \toprule
    \multirow{2}{*}{Method} & \multicolumn{3}{|c|}{Full-view} & \multicolumn{3}{c}{Partial-view} \\
     % & \multicolumn{1}{|c}{mIoU} & \multicolumn{1}{c}{Acc} & \multicolumn{1}{c|}{mAcc} & \multicolumn{1}{c}{mIoU} & \multicolumn{1}{c}{Acc} & \multicolumn{1}{c}{mAcc} \\
    & \multicolumn{1}{|c}{$\mathrm{mIoU}^\mathrm{c}$} & \multicolumn{1}{c}{$\mathrm{Acc}^\mathrm{c}$} & \multicolumn{1}{c|}{$\mathrm{mAcc}^\mathrm{c}$} & \multicolumn{1}{c}{$\mathrm{mIoU}^\mathrm{c}$} & \multicolumn{1}{c}{$\mathrm{Acc}^\mathrm{c}$} & \multicolumn{1}{c}{$\mathrm{mAcc}^\mathrm{c}$} \\
    \midrule   
    \multicolumn{1}{l|}{\gray{Point Transformer~\citep{zhao2021point_transformer}}}
    &\multicolumn{1}{c}{\gray{41.26}} & \multicolumn{1}{c}{--} &\multicolumn{1}{c|}{\gray{67.03}} &\multicolumn{1}{c}{\gray{40.51}} & \multicolumn{1}{c}{--} &\multicolumn{1}{c}{\gray{65.34}}  \\
    
    \multicolumn{1}{l|}{\gray{PointNet++~\citep{qi2017pointnet++}}}
    &\multicolumn{1}{c}{\gray{41.26}} & \multicolumn{1}{c}{--} &\multicolumn{1}{c|}{\gray{\textbf{68.14}}} &\multicolumn{1}{c}{\gray{41.10}} & \multicolumn{1}{c}{--} &\multicolumn{1}{c}{\gray{\textbf{66.74}}}  \\
    
    \multicolumn{1}{l|}{\gray{DGCNN~\citep{wang2019dynamic}}}
    &\multicolumn{1}{c}{\gray{42.09}} & \multicolumn{1}{c}{--} &\multicolumn{1}{c|}{\gray{61.47}} &\multicolumn{1}{c}{\gray{41.93}} & \multicolumn{1}{c}{--} &\multicolumn{1}{c}{\gray{63.12}}  \\
    \midrule    
     
    \multicolumn{1}{l|}{OpenAD-PointNet++~\citep{nguyen2023open}} 
    & \multicolumn{1}{c}{40.17} & \multicolumn{1}{c}{38.61} & \multicolumn{1}{c|}{66.83} 
    & \multicolumn{1}{c}{40.44} & \multicolumn{1}{c}{38.92} & \multicolumn{1}{c}{65.84} \\
    
    \multicolumn{1}{l|}{OpenAD-DGCNN~\citep{nguyen2023open}} 
    & \multicolumn{1}{c}{41.17} & \multicolumn{1}{c}{35.71} & \multicolumn{1}{c|}{59.17} 
    & \multicolumn{1}{c}{39.87} & \multicolumn{1}{c}{35.15} & \multicolumn{1}{c}{59.27} \\
        
    \multicolumn{1}{l|}{IAGNet~\citep{yang2023IAGNet}} 
    & \multicolumn{1}{c}{40.04} & \multicolumn{1}{c}{35.12} & \multicolumn{1}{c|}{53.05} 
    & \multicolumn{1}{c}{41.24} & \multicolumn{1}{c}{34.68} & \multicolumn{1}{c}{52.58} \\
    
    \multicolumn{1}{l|}{LASO~\citep{li2024laso}} 
    & \multicolumn{1}{c}{41.31} & \multicolumn{1}{c}{35.02} & \multicolumn{1}{c|}{53.96} 
    & \multicolumn{1}{c}{40.11} & \multicolumn{1}{c}{35.21} & \multicolumn{1}{c}{52.68} \\    

   \midrule    
    \multicolumn{1}{l|}{3D-ADLLM} 
    & \multicolumn{1}{c}{\textbf{42.85}} & \multicolumn{1}{c}{\textbf{41.84}} & \multicolumn{1}{c|}{66.35} 
    & \multicolumn{1}{c}{\textbf{41.92}} & \multicolumn{1}{c}{\textbf{43.40}} & \multicolumn{1}{c}{61.93} \\
    
    \bottomrule
    \end{tabular}%
    }
\end{table}%
\begin{table}[htbp]
\caption{The performance of close set affordance detection over \textbf{all instances}.}
\label{appendix: close-set task result over all instances}
\begin{center}
\renewcommand{\arraystretch}{1.1} % 增加行高
\begin{tabular}{c | l c c c c c}
\toprule
\rotatebox{90}{} 
% & Method & mIoU  & mAcc & mPrec & mRec & mAP$_{50}$\\
& Method & $\mathrm{mIoU}^\mathrm{i}$  & $\mathrm{mAcc}^\mathrm{i}$ & $\mathrm{mPrec}^\mathrm{i}$ & $\mathrm{mRec}^\mathrm{i}$ & $\mathrm{mAP}_{50}^\mathrm{i}$\\
\midrule
\multirow{4}{*}{\rotatebox{90}{Full-view}} 
& OpenAD-PointNet++  & {28.34} & {64.11} & {33.91} & {61.45} & {5.12} \\
& OpenAD-DGCNN      & {26.98} & {65.94} & {34.38} & {54.76} & {4.77} \\
& LASO               & {44.43} & \textbf{{83.80}} & \textbf{{62.73}} & {60.25} & {21.13} \\
& 3D-ADLLM (ours)    & \textbf{46.29} &81.24  & 57.90 & \textbf{64.27} & \textbf{46.38} \\
\midrule
\multirow{4}{*}{\rotatebox{90}{Partial-view}} 
& OpenAD-PointNet++ & {29.50} & {63.26} & {35.21} & {61.34} & {6.77} \\
& OpenAD-DGCNN      & {17.07} & {67.11} & {27.96} & {30.15} & {1.87} \\
& LASO              & {43.35} & {\textbf{82.31}} & {60.27} & {59.57} & {20.85} \\
& 3D-ADLLM (ours)   & \textbf{44.06} & 79.64 & 56.23 & \textbf{64.21} & \textbf{46.60} \\
\bottomrule
\end{tabular}
\end{center}
\end{table}

\subsection{Data Analysis}
\label{apendix:data analysis}
\subsubsection{The detailed settings of full view and partial view.}
Building the IRAS dataset based on the 3D AffordanceNet (OpenAD) Dataset~\citep{nguyen2023open}. We retained the settings for both full-view point cloud and partial-view point cloud.

Full-view: Given an object as 3D point cloud without knowing the affordances supported by the object, the full-shape affordance estimation task aims to estimate the supported affordance type and predict the point-wise probabilistic score of affordance.

Partial-view: in real-world application scenarios, we can only expect partial view of 3D shapes, represented as partial point cloud. Therefore, another important task we are concerned with is to estimate the affordance from partial point cloud.
\subsubsection{The visualization of Data Analysis.}
\begin{figure}[htbp]
    \centering
    \includegraphics[width=1.0\linewidth]{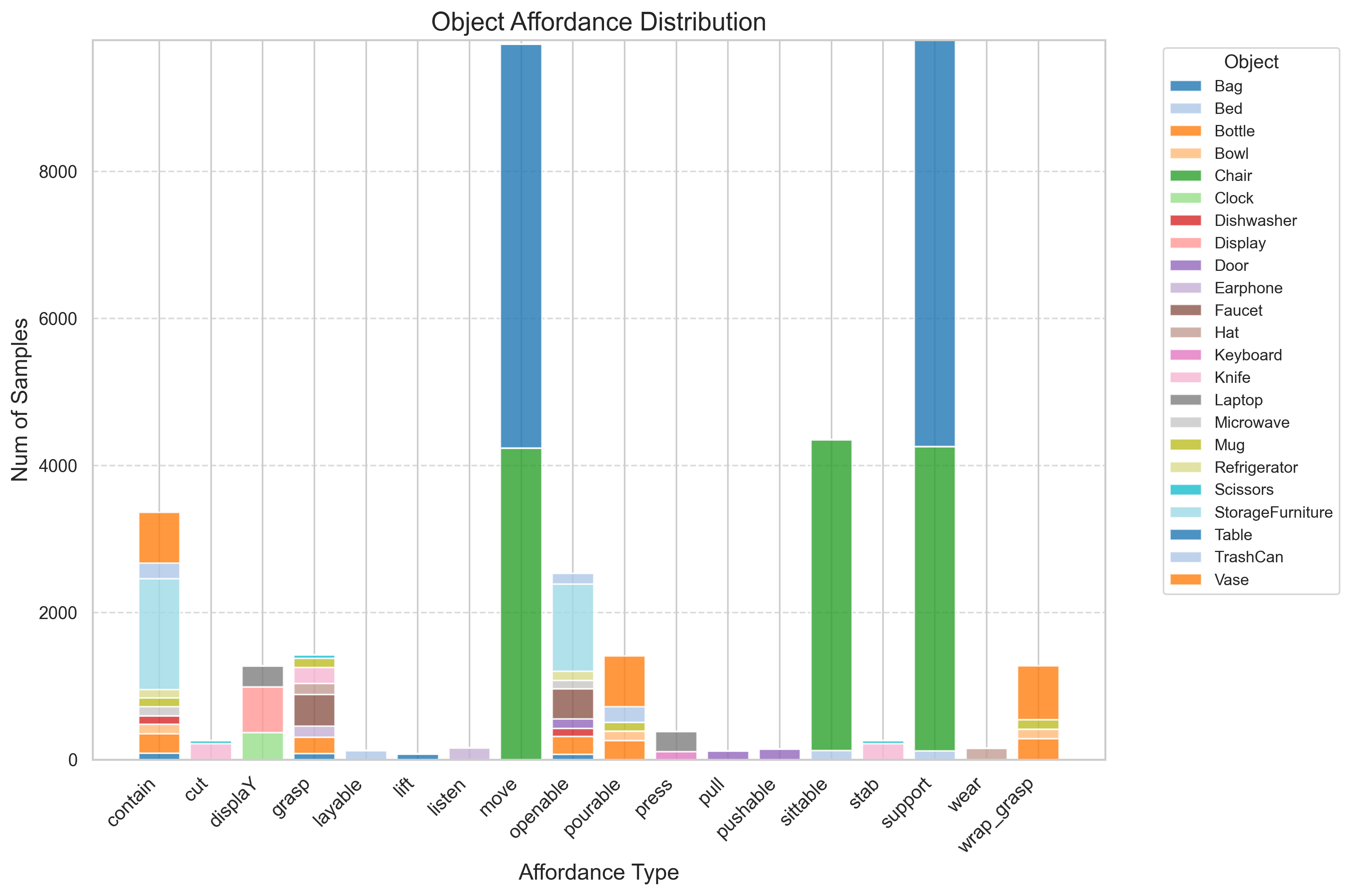}
    \caption{The analysis of IRAS task.}
    \label{fig:3D AffordanceNet dataset}
\end{figure}
\begin{figure}
    \centering
    \includegraphics[width=1\linewidth]{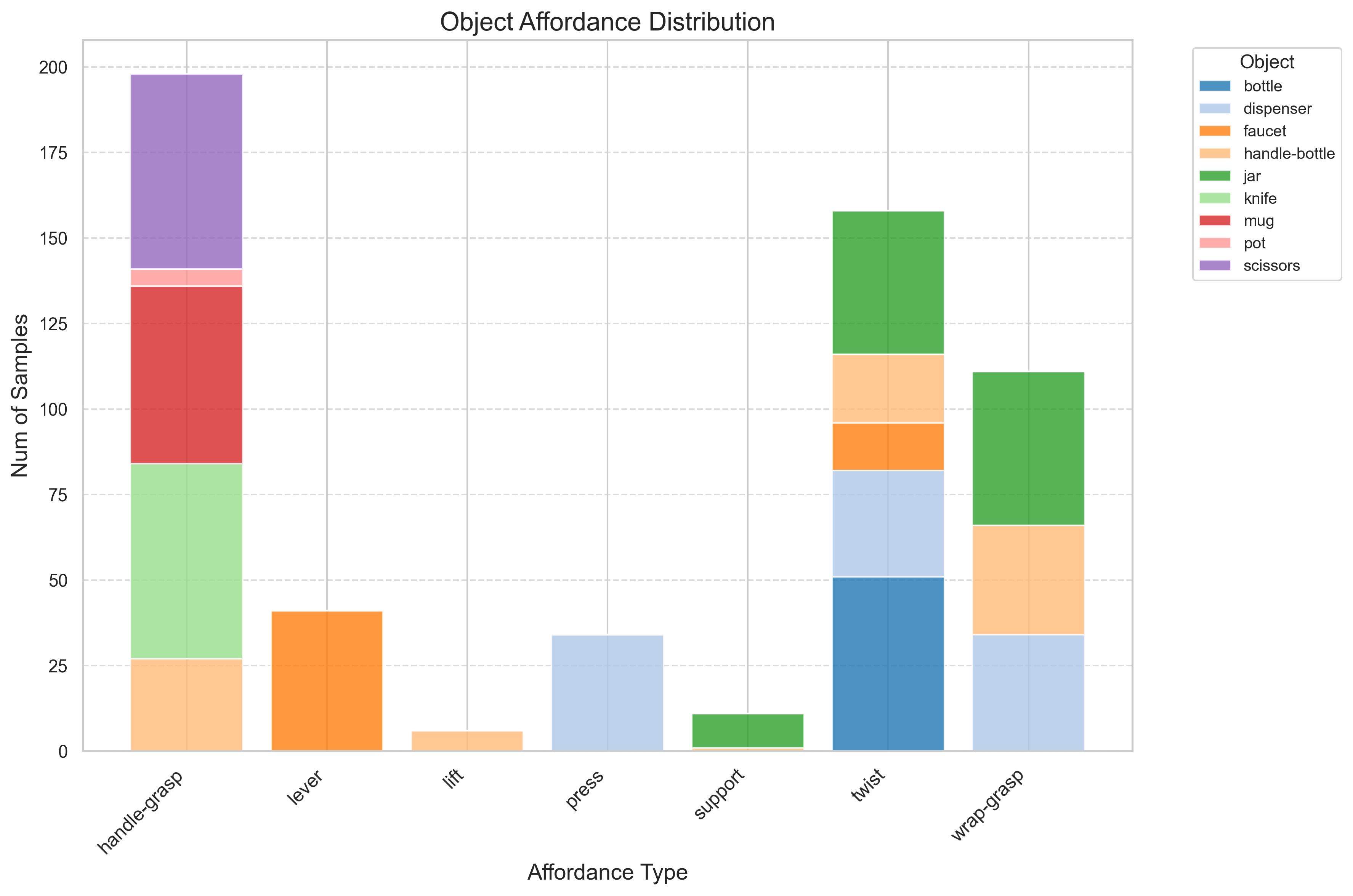}
    \caption{The analysis of extensive test dataset in sec.~\ref{sec:out of distribution result}.}
    \label{fig:AffordPose}
\end{figure}
\begin{figure}
    \centering
    \includegraphics[width=1\linewidth]{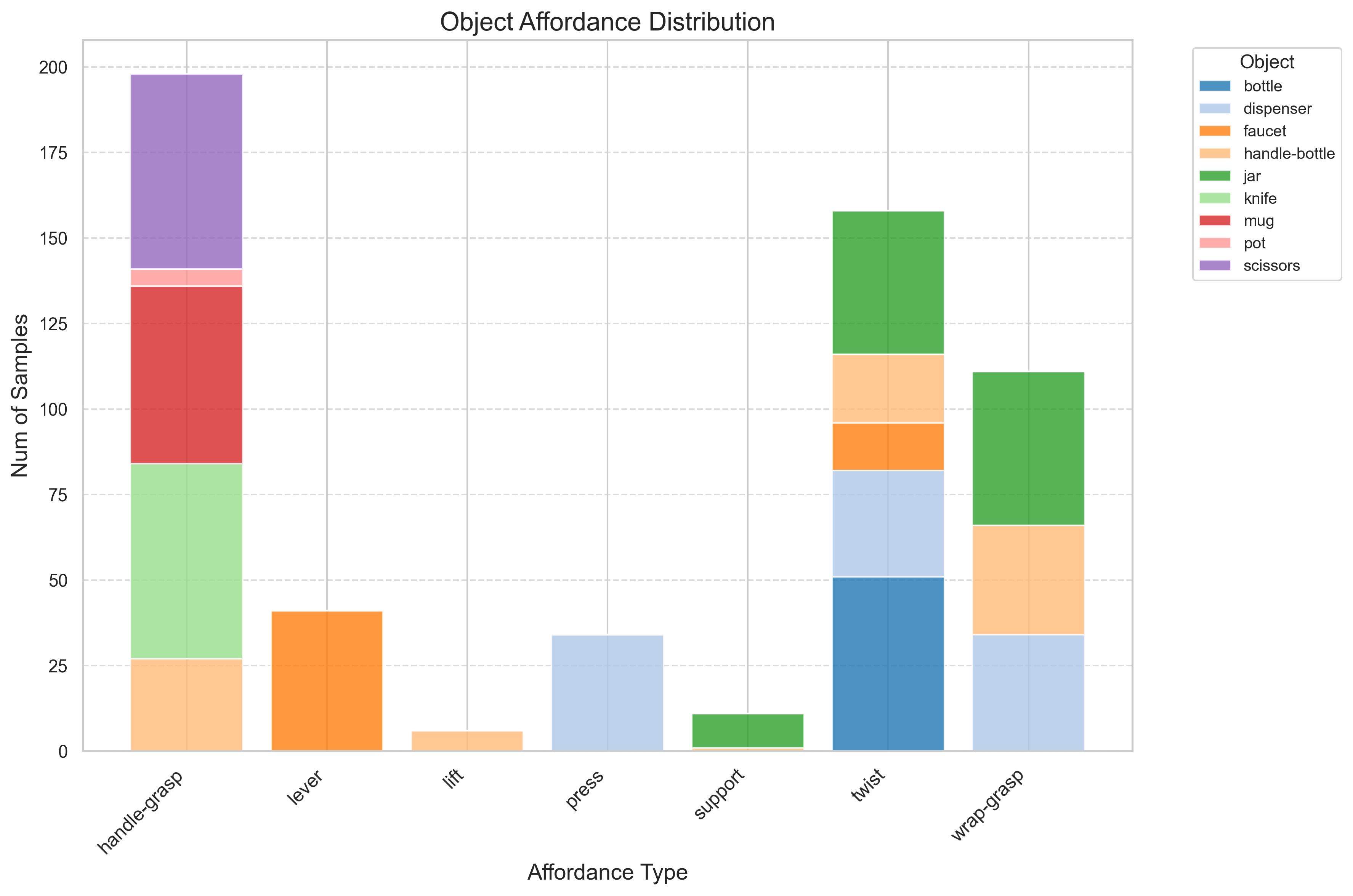}
    \caption{The analysis of ROPS task.}
    \label{fig:Partnet}
\end{figure}

\subsection{Training details}
We constructed our IROS dataset based on the PartNet dataset and aim to acquire general segmentation knowledge through the pre-training phase. To balance training costs with model performance, we selectively sampled a subset of the data based on categories to obtain general segmentation knowledge for objects.

\end{document}